\documentclass[11pt,draftcls,onecolumn]{IEEEtran}
\usepackage[section]{algorithm}
\usepackage{algorithmic}
\usepackage{array}
\usepackage{amsmath}
\usepackage{amssymb}
\usepackage{amsfonts}
\usepackage{amsthm}
\usepackage{booktabs}
\usepackage{color}
\usepackage{graphicx}
\usepackage{subfigure}
\usepackage{multirow}
\usepackage{threeparttable}
\usepackage{hyperref}
\usepackage{makecell}
\usepackage{fmtcount}
\usepackage{colortbl}

\numberwithin{equation}{section}

\theoremstyle{remark}

\newcommand{\eg}{e.g.}

\newcommand{\cC}{\mathcal{C}}

\newcommand{\cN}{\mathcal{N}}

\newcommand{\R}{\mathbb{R}}

\newcommand{\eps}{\varepsilon}

\title{Higher-order MRFs based image super resolution: why not MAP?}
\author{Yunjin Chen \thanks{Y.J. Chen is with the Institute for Computer Graphics and Vision, Graz 
University of Technology, Inffeldgasse 16, A-8010 Graz, Austria. e-mail: 
\href{mailto:chenyunjin\_nudt@hotmail.com}{chenyunjin\_nudt@hotmail.com}. 

This work was supported by the Austrian Science Fund (FWF) under the START project
BIVISION, No. Y729.
}}

\begin{document}

\markboth{IET Image Processing,~Vol.~xx, No.~xx, 2015}
{Chen: Higher-order MRFs based SR: why not MAP?}
\maketitle

\begin{abstract}
A trainable filter-based higher-order Markov Random Fields (MRFs) model - the so called Fields of Experts (FoE), 
has proved a highly effective image prior model for many classic image restoration problems. Generally, two options are 
available to incorporate the learned FoE prior in the inference procedure: (1) sampling-based 
minimum mean square error (MMSE) estimate, and (2) energy minimization-based maximum a posteriori (MAP) estimate. 
This paper is devoted to the FoE prior based single image super resolution (SR) problem, and 
we suggest to make use of the MAP estimate for inference based on two facts: (I) It is well-known that 
the MAP inference has a remarkable advantage of high computational efficiency, while the sampling-based MMSE estimate 
is very time consuming. (II) Practical SR experiment results demonstrate that the MAP estimate works equally well compared 
to the MMSE estimate with exactly the same FoE prior model. Moreover, it can lead to even further improvements 
by incorporating our discriminatively trained FoE prior model. 
In summary, we hold that for higher-order natural image prior based SR problem, it is better to employ the MAP estimate for 
inference. 
\end{abstract}


\IEEEpeerreviewmaketitle

\section{Introduction}
Markov Random Fields (MRFs) based models have a long history in low-level 
computer vision problems, which treat the image (can be also seen as a 
label field) as a random field \cite{li1995markov}. 
It is well-known that MRFs are particularly effective for image/label prior modeling 
in image processing. In a MRF-based prior model, the probability 
of a whole field is defined based on the potential (or energy) of the local cliques. 
{MRF-based prior models, especially higher-order potentials 
for enforcing label consistency have demonstrated great successes in the community of 
multi-label image segmentation \cite{nguyen2012higher, kohli2009robust}. The exploited 
higher-order potentials are constructed on sets of pixels (i.e., super-pixels) in 
contrast to conventional pair-wise cliques defined on a 4 or 8 neighborhood. }

Recently, an elegant MRF-based prior model specialized for natural image, 
called Fields of Experts (FoE) was proposed by Roth and Black \cite{RothFOE2009}. 
The proposed FoE model is defined by (1) a heavy-tailed potential 
function, which is derived from the observation that the filter response of 
natural images exhibit heavy-tailed distribution when applying 
derivative filters onto them, (2) a set of linear filters, which are 
trained from image samples. {
The FoE model has a larger clique structure that is capable of capturing higher order
interactions around each image pixel than models based on pairwise interactions.
The FoE model differs from those MRF models, such as 
\cite{nguyen2012higher, kohli2009robust} in three aspects: (a) it is a 
continuously-valued MRF model in contrast to a discrete field with finite labels; 
(b) its local potentials are built on the filter response of certain linear 
filters; (c) its connections related to neighborhood labels are 
directly based on image pixels, not super-pixels. }

Due to its effectiveness of the FoE image prior model for many 
image restoration problems, 
many works have been devoted to the FoE-based image restoration problems, such as image 
denoising, inpainting, deblurring, etc \cite{RothFOE2009, SamuelFoE, ChenPRB13}. 
Usually, there are two ways to investigate the learned FoE prior model for specific image restoration problems, 
the sampling-based MMSE estimation, such as \cite{gaocrpr2010, GaoDAGM2012, zhang2012bayesian, zhang2012generative}, 
and the energy minimization based MAP estimation, such as \cite{RothFOE2009, SamuelFoE, ChenPRB13, ChenRP14}. 

It is \cite{gaocrpr2010} that for the first time claimed that the MMSE 
estimation can lead to better performance compared to the MAP estimation for the image denoising task with their learned FoE 
image prior model. 
After that, many works follow their suggestion to make use of the MMSE estimation for FoE related models, such as 
image deblurring \cite{schmidt2011bayesian, zhao2013non}, image denoising \cite{GaoDAGM2012}, depth estimation 
\cite{wang2014depth, herrera2013learned}, image separation \cite{zhang2012bayesian} and single image super resolution 
\cite{zhang2012generative}. 

In a recent paper \cite{zhang2012generative}, the FoE prior model was exploited in the context of image super
resolution. The authors also proposed to employ the MMSE estimate in the inference procedure 
instead of the MAP estimate. With the MMSE 
estimate, the FoE-based SR model demonstrates a state-of-the-art SR algorithm. 
However, it is well known that the sampling based approach is very time consuming, 
alluding to the fact that the FoE-based SR model is not appealing for practical applications. 

It is generally true that the MMSE estimate is a better alternative than MAP, 
as it can exploit the uncertainty of the model, especially in the case of 
multimodal distribution with multiple peaks. However, in practice it is usually 
intractable to find an accurate solution for the 
MMSE estimate due to the difficulty of taking the expectations over entire images. 
{Therefore, approximate approaches, 
such as the commonly used sampling based algorithm is considered. 
This gives the MAP inference, which seeks the maximum
peak, a chance to work equally well.}

In this paper, we evaluate the performance of the MAP inference for the FoE based SR problem. Our experimental results 
demonstrate that the MAP inference of the FoE-based SR model has been 
underestimated in the previous work \cite{zhang2012generative}. Numerical results show that 
with exactly the same image prior model exploited in the MMSE estimation, the MAP inference can achieve equivalent performance 
in terms of both quantitative measurements (PSNR and SSIM values) and visual perception quality. 
In addition, the MAP inference can obtain further improvements with the discriminatively 
trained FoE image prior of the same model capacity. 
It is clear that the MAP inference has a significant advantage of efficiency, and this advantage is even more remarkable 
with our recently proposed non-convex optimization algorithm - iPiano \cite{ipiano}. 

To sum up, our experimental findings suggest us 
to exploit the MAP inference for solving the FoE prior-based image super resolution problem, 
because (1) there is no performance loss by using this simpler inference criterion, 
and (2) the MAP inference has an apparent advantage of high efficiency. 

\section{MAP inference of FoE image prior based SR}
In a typical image super resolution task, the low-resolution (LR) image is generated from a high-resolution (HR) image using 
the following formulation
\[
y = DBx + \eps\,,
\]
where $x \in \R^n$ and $y \in \R^m$ is the HR and LR image, respectively. $B \in \R^{n \times n}$ is the matrix corresponding to 
the blurring operation and $D \in \R^{m \times n}$ ($m < n$) signifies the down-sampling operation. $\eps \in \R^m$ is the noise 
(typically assumed to be Gaussian white noise with level $\sigma$).

The FoE image prior based SR model is formulated by the following Bayesian probabilistic model
\begin{equation}\label{posterior}
p(x|y) \propto \text{exp}(-\frac {\|y - DBx\|_2^2}{\sigma^2})p(x) \,,
\end{equation}
where $p(x)$ is the probability density of an image $x$ under the FoE framework, written as 
\[
p(x) = \frac 1 Z \prod\limits_{c \in \cC}\prod\limits_{i=1}^{N} \phi((k_i*x)_c; \alpha_i)
\]
where $\cC$ is the maximal cliques, $N$ is the number of the filters, 
$(k_i*x)_c$ refers to the $c$-th pixel in the filtered image by $k_i$, $\phi$ is the potential function 
with associated weights $\alpha$. In \cite{zhang2012generative}, 
the potential function is given by the Gaussian scale mixtures (GSMs) as
\begin{equation}\label{potential}
\phi(z; \alpha_i) = \sum\limits_{j=1}^{J}\alpha_{i,j} \cN(z; 0, \eta_i^2/s_j)\,,
\end{equation}
where $\alpha_{i,j}$ are the normalized weights of the Gaussian component with scale $s_j$ and base variance $\eta_i^2$. 

According to the posterior \eqref{posterior}, \cite{zhang2012generative} 
used the sampling-based 
MMSE estimation to recover the underlying HR image $x$. 
{The MMSE estimate is given by the following problem
\begin{equation}\label{mmse}
\hat x = \arg \min\limits_{x}\int \|x - \hat x\|_2^2 p(x|y)dx = E[x|y] \,.
\end{equation}
MMSE is equal to the mean of the posterior distribution
and generally differs from the maximum (i.e, MAP) in case of non-
Gaussian posteriors, as exploited in this paper. 
As shown in \cite{zhang2012generative, gaocrpr2010}, it is typically intractable to 
solve \eqref{mmse}, due to the difficulty of taking expectations 
over entire images. Usually, an approximate approach based on 
samples drawn from the posterior distribution is used. 
}

In this paper, we consider the MAP estimate. 
With the MAP estimation, the FoE-based SR task is formulated as the following energy minimization problem 
\begin{equation}\label{MAP}
\hat x = \arg\min\limits_{x} E(x) = \sum\limits_{i=1}^{N}\rho(k_i*x) + \frac \lambda 2 \|DBx - y\|_2^2 \,,
\end{equation}
where $\rho(k_i*x) = \sum\nolimits_{c \in \cC} \rho((k_i*x)_c)$ with penalty 
function $\rho = -\text{log} \phi$ defined in \eqref{potential}. { 
Note that the penalty function $\rho$ is a non-convex function, as the potential 
function $\phi$ is heavy-tailed.}

In our work, we consider a newly developed non-convex optimization - iPiano \cite{ipiano} to solve the above minimization 
problem, instead of the commonly used conjugate 
gradient (CG) algorithm. We find that the iPiano  algorithm is significantly faster than CG. 
We refer the interested readers to \cite{ipiano} for more details about the iPiano algorithm. 

\section{Experimental results}
We mainly conducted two types of experiments. The first type is to perform a direct comparison between the MAP estimate and 
the MMSE estimate for the FoE based SR task. The second type is to compare the MAP based SR model 
to very recent state-of-the-art SR approaches. The corresponding implementations are all from publicly available codes 
provided by the authors, and are used as is. 
{
\subsection{Solving the corresponding non-convex optimization problems via iPiano}
As mentioned before, we make use of the iPiano algorithm to solve the 
non-convex problem \eqref{MAP}. 
The iPiano algorithm is designed for solving an optimization 
problem which is composed of
a smooth (possibly non-convex) function $F$ and a convex (possibly non-smooth) 
function $H$:
\begin{equation}\label{fplusg}
\arg\min\limits_{x}F(x) + H(x) \,.
\end{equation}
It is based on a forward-backward splitting scheme with an inertial force term. 
Its basic update rule is given as
\begin{equation}\label{iPianoupdate}
x^{n+1}=\left( I+\tau \partial H \right)^{-1}\left( x^n-\tau \nabla F\left( x^n \right) +
\gamma \left( x^n - x^{n-1} \right) \right),
\end{equation}
where $\tau$ and $\gamma$ are the step size parameters. The term $\left( I+\tau \partial H \right)^{-1}$ signifies the
standard proximal mapping \cite{nesterov2004introductory}, which is the backward step.
$x^n-\tau \nabla F\left( x^n \right)$ is the forward gradient descent step,
and $\gamma \left( x^n - x^{n-1} \right)$ is the inertial term.

Casting \eqref{MAP} in the form of \eqref{fplusg}, we set 
$F(x) = \sum\limits_{i=1}^{N}\rho(k_i*x) + \frac \lambda 2 \|DBx - y\|_2^2$ and 
$H(x) = 0$. It is easy to check that gradient $\nabla_x F$ is given as 
\begin{equation}\label{gradient}
\nabla_x F = \sum\limits_{i=1}^{N}K_i^\top\rho'(K_ix) + \lambda (DB)^\top(DBx - y) \,,
\end{equation}
where $K_i \in\R^{n \times n}$ a highly sparse matrix, implemented as 2D convolution of the image $x$ with filter kernel $k_i$, 
i.e., $K_ix \Leftrightarrow k_i*x$, $\rho'(K_i x) = (\rho'((K_i x)_1),\cdots,\rho'((K_i x)_n))^\top \in \R^{n}$, with 
$\rho'(z) = \frac{z}{\phi(z)}\sum\limits_{j=1}^{J}\frac{s_j\alpha_{i,j}}{\eta_i^2} \cN(z; 0, \eta_i^2/s_j)$. The proximal mapping operation 
with respect to $H$ is given as 
\begin{equation}\label{solution-subproblemG}
\left( I+\tau \partial H \right)^{-1}(\hat x) = \hat x \,.
\end{equation}

Note that we can also consider the setting that 
$F(x) = \sum\limits_{i=1}^{N}\rho(k_i*x)$ and $H(x) = \frac \lambda 2 \|DBx - y\|_2^2$. 
Then, the proximal mapping operation with respect to $H$ is given as 
\begin{equation}\label{solution-subproblemG}
\left( I+\tau \partial H \right)^{-1}(\hat x) = \min\limits_{x}
\frac{\|x - \hat x\|_2^2}{2} +  \frac {\tau\lambda }{2} \|DBx - y\|_2^2 
= \left(I + \tau\lambda (DB)^\top DB\right)^{-1}
\left(\hat x + \tau\lambda(DB)^\top y\right)\,.
\end{equation}
As this sub-problem involves an inverse matrix computation, which is time 
consuming in practice, we prefer the former setting. 
Therefore, the overall process for MAP based SR by using the iPiano algorithm is 
summarized in Algorithm \ref{algo2}. 

\begin{algorithm}\caption{The overall MAP-based SR process using the iPiano algorithm
}\label{algo2}
\textbf{Input}: The observed LR image $y$ \\
\textbf{Initialization}:set \bfseries{Iter}$>0$,
\rm{Choose} $\gamma = 0.8$, $l_{-1} = 1$, $\eta = 1.2$,
and initialize $x^0$ using bi-cubic interpolation and set $x^{-1}=x^0$,
\bfseries{For}\mdseries { }$n$ = 0:$(\textbf{Iter}-1)$
\begin{itemize}
\item[1.] Conduct a line search to find the smallest
nonnegative
integer $i$ such that with $l_n=\eta^i l_{n-1}$,
\begin{equation}\label{ldet}
F ( x^{n+1}) \leq  F\left( x^{n} \right) + \langle \nabla F\left( x^n \right),x^{n+1}-x^n \rangle \nonumber +\frac{l_n}{2} \| x^{n+1}-x^n \|^2_2.
\end{equation} 
is satisfied;
\item[2.] Set $l_n=\eta^i l_{n-1}$, $\tau_n = 1.99\left( 1-\gamma\right)/l_n$;
\item[3.] Compute $\nabla_x F\left( x^n \right)$ according to \eqref{gradient};
\item[4.] Compute
$x^{n+1}$ according to \eqref{iPianoupdate} and \eqref{solution-subproblemG};\\
\textbf{end For}\\
$\hat{x}={x^{\mathbf{Iter}}}$;
\end{itemize}
\textbf{Solution}: Estimated underlying HR image $\hat x$.
\end{algorithm}

}

\subsection{Comparison between the MAP and MMSE estimate}
In order to conduct a fair comparison with the MMSE estimation, we first considered the MAP estimation 
with exactly the same image prior model exploited in \cite{zhang2012generative} 
(8 filters of size $3 \times 3$ with GSMs potential). 
We repeated the experiments presented in the TABLE I of \cite{zhang2012generative}, where eight noise-free images 
were upsampled with a zooming factor of 3. The results of the MMSE and MAP estimates are shown in Table \ref{table:noisefree}. 
One can see that the MAP estimate using the same image prior model 
performs equally well compared to the MMSE estimate, in terms of PSNR and SSIM index\footnote{
Note that we were not able exactly reproduce the results presented in \cite{zhang2012generative} due to the 
randomness of the sampling-based approach. We actually achieved slightly different results. }.

\begin{table*}[t!]
\centering
\hspace*{-1cm}\begin{tabular}{l|c c c c c c c c}
\hline
& House & Peppers & Cameraman & Barbara & Lena & Boat & Hill & Couple\\
\hline \hline
MMSE with prior \eqref{potential} & 31.73/88.85 & \color{blue}25.94/90.94 &\color{blue}26.26/83.43 &25.55/74.44 &32.93/90.34 
&{\color{blue}29.32}/83.32 &30.28/81.83 &{\color{blue}28.47}/80.34 \\
MAP with prior \eqref{potential} & \color{blue}32.25/89.03 & 25.86/89.50 &25.91/82.20 &\color{blue}25.65/75.41 
&\color{blue}33.16/90.97 &29.10/\color{blue}83.53 &\color{blue}30.71/82.59 &28.41/\color{blue}80.55 \\
\Xhline{0.5pt}
\hline
MAP with prior \eqref{stFOE}& \textbf{32.72/89.61} & \textbf{26.62/91.43} &\textbf{26.69/84.66} &\textbf{25.71/75.71} 
&\textbf{33.52/91.44} &\textbf{29.48/84.47} &\textbf{31.14/83.77} &\textbf{28.77/81.91} \\
\hline
\end{tabular}
\vspace{0.2cm}
\caption{SR ($\times 3$) result comparison between the MMSE and MAP estimate (\textbf{PSNR/$100 \times$SSIM}). 
Better results of the first two rows (with the same FoE prior) are colored with {\color{blue}blue}. 
The best results are highlighted in bold.}\label{table:noisefree}
\vspace{-0.2cm}
\end{table*}

\begin{figure}[t!]
\vspace*{-0.5cm}
\centering
    \subfigure[]{\includegraphics[width=0.4\textwidth]{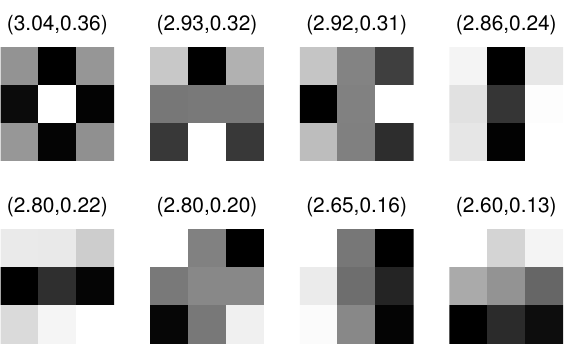}}
    \subfigure[]{\includegraphics[width=0.3\textwidth]{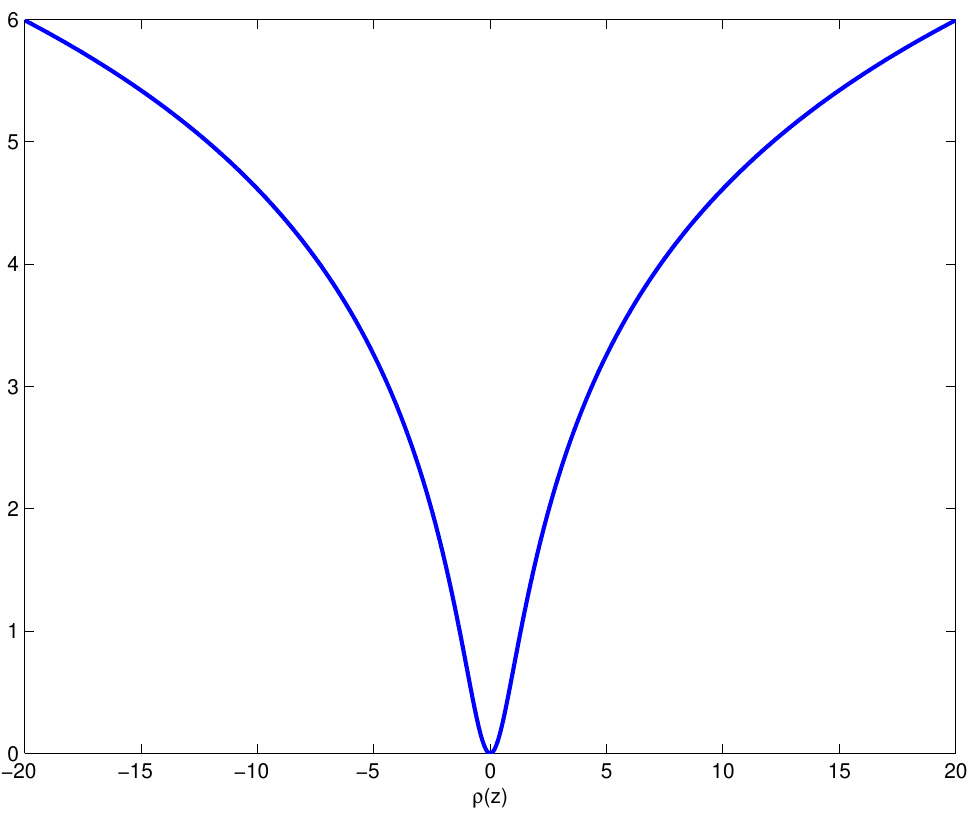}}\\
    \caption{Discriminatively trained FoE prior with the Lorentzian penalty function. (a) the learned filters. 
The first number in the bracket is weight $\theta_i$ and the second one is the norm of the filter $k_i$. (b) the corresponding 
Lorentzian penalty function derived from the Student-t distribution.}\label{filters}
\vspace*{-0.25cm}
\end{figure}

We then exploited a discriminatively trained FoE prior for the MAP-based SR model to further 
investigate its performance. The discriminatively trained FoE prior has the same model capacity, and 
is directly optimized based on the MAP estimate in 
the context of Gaussian denoising. We employed the Student-t based FoE model trained in our previous work \cite{ChenRP14}, 
which is defined as 
\begin{equation}\label{stFOE}
E_{FoE_{st}}(x) = \sum\limits_{i=1}^{N}\theta_i \rho(k_i*x) \,,
\end{equation}
where the penalty function is given as the Lorentzian function $\rho(z) = \text{log}(1+z^2)$ 
shown in Figure \ref{filters}(b), and $\theta_i$ is the weight of the corresponding filter $k_i$. 
The corresponding filters are shown in Figure \ref{filters}(a). 

\begin{figure}[t!]
\vspace*{-0.1cm}
\centering
    \subfigure[Original ($510 \times 510$)]{\includegraphics[width=0.33\textwidth]{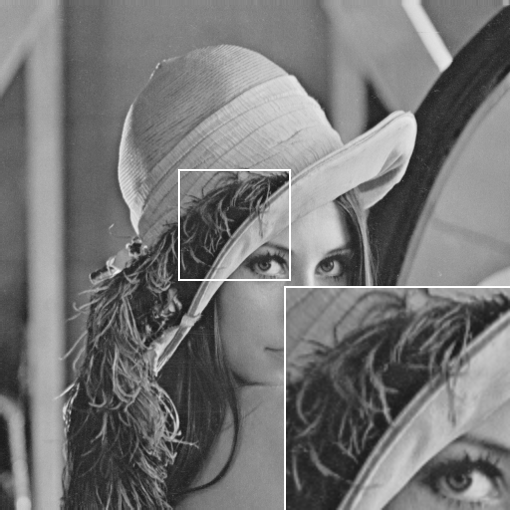}}\hfill
    \subfigure[Low resolution input ($170 \times 170$)]{\includegraphics[width=0.33\textwidth]{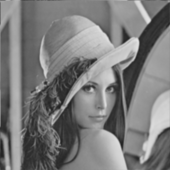}}\hfill
    \subfigure[bicubic interpolation (26.62/80.20)]{\includegraphics[width=0.33\textwidth]{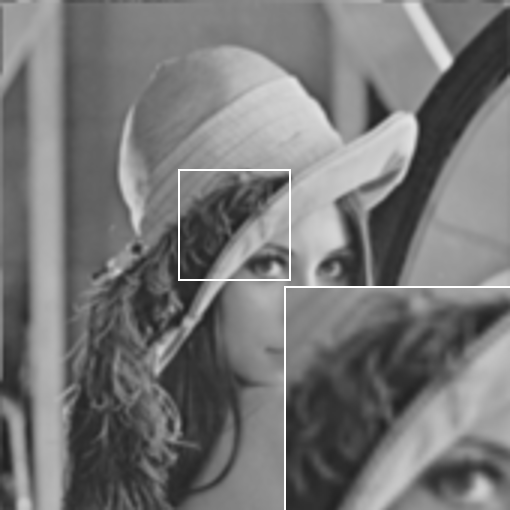}}\\
    \subfigure[MMSE with \eqref{potential} (32.93/90.34)]{\includegraphics[width=0.33\textwidth]{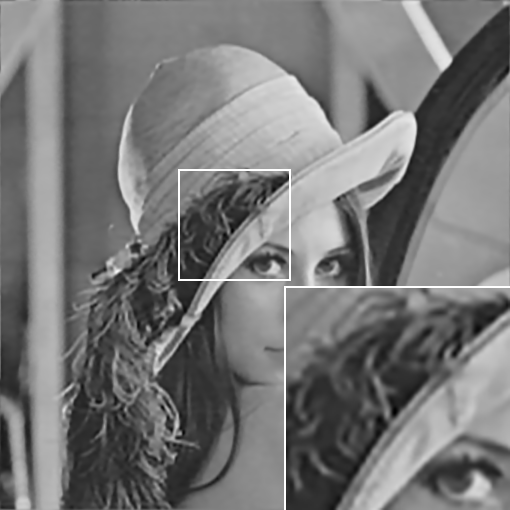}}\hfill
    \subfigure[MAP with \eqref{potential} (33.16/90.97)]{\includegraphics[width=0.33\textwidth]{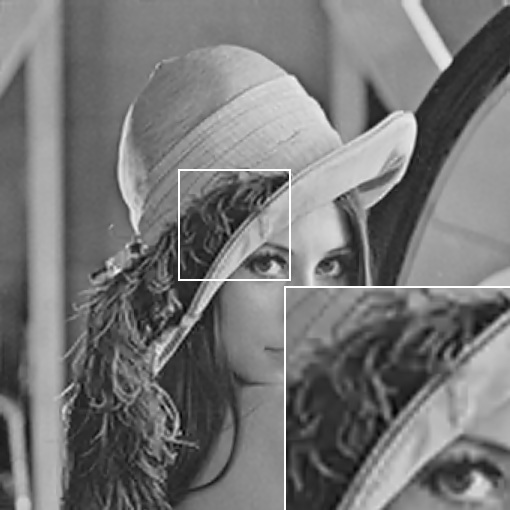}}\hfill
    \subfigure[MAP with \eqref{stFOE} (\textbf{33.52/91.44})]{\includegraphics[width=0.33\textwidth]{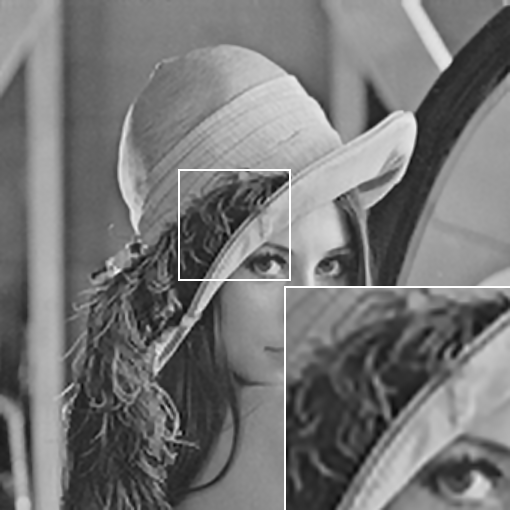}}\\
    \caption{Single image super resolution results for the noise-free ``Lena'' image ($\times 3$) with different algorithms. The results are 
evaluated in terms of PSNR and SSIM ($\times 100$) index.}\label{lena}
\end{figure}

The results of the MAP-based SR model with this discriminatively 
trained FoE prior \eqref{stFOE} are also shown in 
Table \ref{table:noisefree}. One can see that the MAP inference 
with our discriminatively trained FoE model improves 
the PSNR and SSIM results. An illustrative example is presented in Figure \ref{lena}.

{
It is not surprising that the discriminatively trained FoE model is able to 
improve the results of the generative FoE model exploited in 
\cite{zhang2012generative}, by using the MAP inference. The reason reads as follows: 
if our intent is to use MAP estimate and evaluate the estimator by 
some criteria, such as PSNR, a better strategy is to train the 
parameters (i.e., the FoE prior model) 
such that the performance of MAP estimate is directly optimized 
(i.e., the so called discriminative training), 
in contrast to training in a maximum-likelihood fashion and 
then conducting inference with MAP. }

\begin{table}[t!]
\centering
\begin{tabular}{l|c c c c}
\hline
$\sigma$ & Methods & House & Peppers & Cameraman\\
\hline \hline
\multirow{3}*{1} & MMSE with prior \eqref{potential} & 31.26/\textbf{\color{blue}87.74} & 25.69/\color{blue}88.83 
&\color{blue}26.13/\textbf{82.40}\\
& MAP with prior \eqref{potential} & {\color{blue}31.66}/87.30 & {\color{blue}25.87}/87.93 &25.72/80.91\\ 
\cline{2-5}
& MAP with prior \eqref{stFOE} & \textbf{32.03}/87.55 &\textbf{26.23/89.24} &\textbf{26.17}/81.96\\
\Xhline{0.5pt}
\hline
\multirow{3}*{2} & MMSE with prior \eqref{potential} & 30.47/\color{blue}85.80 & 25.23/\color{blue}86.00 
&\color{blue}25.67/\textbf{80.20}\\
& MAP with prior \eqref{potential} & {\color{blue}30.84}/85.62 & {\color{blue}25.38}/85.75 &25.24/78.82\\ 
\cline{2-5}
& MAP with prior \eqref{stFOE} & \textbf{31.25/85.87} &\textbf{25.49/86.97} &\textbf{25.69}/79.68\\
\Xhline{0.5pt}
\hline
\multirow{3}*{3} & MMSE with prior \eqref{potential} & 29.33/83.21 & 24.54/82.32 &\color{blue}24.94/77.04\\
& MAP with prior \eqref{potential} & \color{blue}30.30/84.55 & \color{blue}24.88/83.80 &24.65/76.53\\ 
\cline{2-5}
& MAP with prior \eqref{stFOE} & \textbf{30.59/84.63} &\textbf{25.10/85.19} &\textbf{25.26/77.97}\\
\hline
\end{tabular}
\vspace{0.2cm}
\caption{Noisy image SR ($\times 3$) result comparison between the MMSE and MAP estimate (\textbf{PSNR/$100 \times$SSIM}). 
Better results of the first two rows (with the same FoE prior) are colored with {\color{blue}blue}. 
The best results are highlighted in bold.}\label{table:noise}
\end{table}

We also evaluated the performance of the MAP inference in the presence of noise. 
For the cases of mild Gaussian noise, the results of the MAP inference with two different FoE image prior models are 
shown in Table \ref{table:noise}, together with the results of the MMSE based model. 
Note that this is a direct comparison to TABLE III of \cite{zhang2012generative}. Again, one 
can see that the MAP estimate with the same FoE model (i.e., \eqref{potential}) works equally well, and it leads to better results 
with our discriminatively trained FoE prior \eqref{stFOE}. 

For the MAP estimate based SR model \eqref{MAP}, we need to search an optimal $\lambda$ for each case. 
For the noise-free image SR task, we use a relative large $\lambda = 200$, and for the SR tasks with Gaussian noise, 
we find the following empirical choice (1) $\lambda = 3, \text{if} ~\sigma = 1$, (2) $\lambda = 1, \text{if} ~\sigma = 2$, and 
(3) $\lambda = 0.5, \text{if} ~\sigma = 3$, generally works well. 

\noindent \textbf{Run time:} We run the inference algorithms on a server with 
Inter(R) Xeon(R) CPU E5-2680 v2 @ 2.80GHz. For the SR task of upsampling an image of size $85 \times 85$ to 
the size of $255 \times 255$, the average computation time per iteration of the MMSE-based algorithm is \textbf{87s}. Typically, the 
MMSE estimate takes 100 iterations, and therefore for this SR task, it requires about \textbf{$\sim$2.4h}, making this approach 
hardly appealing for practical application. 

In contrast, the MAP inference is much faster. The average computation time per iteration of the MAP inference is 
\textbf{0.039s} in the case of the Student-t based FoE prior \eqref{stFOE}\footnote{
With the same model capacity of 8 filters of size $3 \times 3$}. 
Typically, it takes 150$\sim$200 iterations to solve the 
resulting non-convex minimization problem\footnote{Also note that the required iterations is dramatically reduced by using the 
iPiano algorithm, compared to the usual CG algorithm used in previous works, 
such as \cite{RothFOE2009, gaocrpr2010}, where the iterative algorithm has to run $\sim$5000 iterations. 
}. As a consequence, the MAP inference with the Student-t based FoE prior 
is able to accomplish the same SR task in \textbf{7s}, which is dramatically 
faster than the MMSE inference (\textbf{$\sim$2.4h}). 
Implementation will be publicly available at the author's 
homepage (\url{http://www.icg.tugraz.at/Members/Chenyunjin}). 

Moreover, as demonstrated in our previous works \cite{ChenPRB13, ChenRP14}, the MAP inference of the FoE prior based models 
can be easily implemented on GPU for parallel computation, which can generally obtain an approximate 
speedup factor of $40\times$. 

\subsection{Comparison to state-of-the-art SR approaches} 
In order to conduct a comprehensive evaluation for the MAP based SR model, 
we further compared it with current state-of-the-art SR approaches: 
the sparse coding (SC) based approach \cite{SCSR}, 
the K-SVD based method \cite{SRKSVD}, the ANR (Anchored 
Neighborhood Regression) based method \cite{ANR} and deep convolutional 
network based method - SR-CNN \cite{SRCNN}. In order to perform 
a fair comparison with these methods, we strictly obey the same test protocols as in \cite{ANR}. We download the source codes from the authors' websites, 
and use the recommended parameters by the authors. We used the same 
test sets - \textbf{Set14} and \textbf{Set5} 
to evaluate the upscaling factor of 3. 

\begin{figure}[t!]
\vspace*{-0.25cm}
\centering
    {\includegraphics[width=0.75\textwidth]{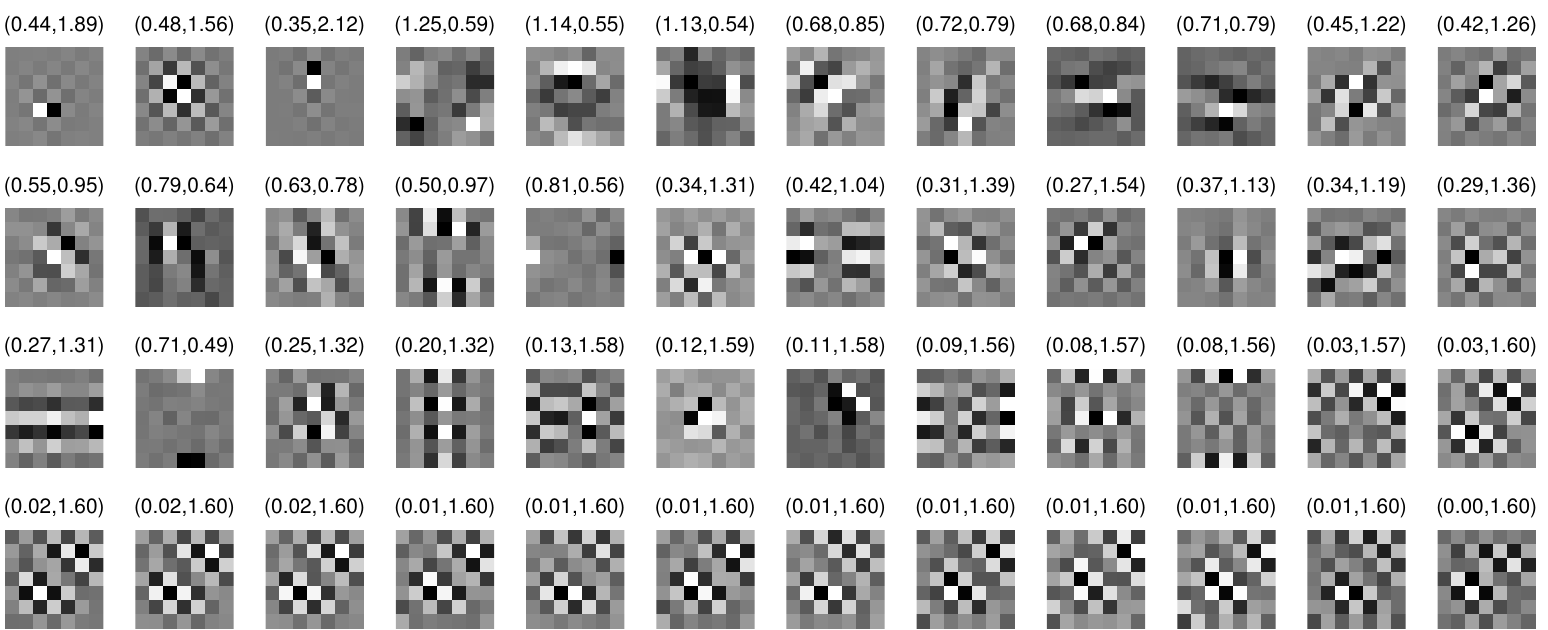}}
\vspace*{-0.3cm}
    \caption{48 learned filters of size $7 \times 7$ exploited in our MAP based SR model.}
\label{filters7x7}
\end{figure}

For the MAP based SR model, we incorporated a FoE prior model with 
larger filter size and more filters (shown in Figure \ref{filters7x7}, 48 filters of size $7 \times 7$), which is trained in \cite{ChenRP14}. 
Replacing the FoE prior model shown in Figure \ref{filters} with 
this new FoE model having increased model capacity can improve the 
performance of the MAP based SR model. 

\begin{table}[t!]
\centering
\begin{tabular}{|l|c| c| c| c| c| c|}
\hline
\textbf{Set14} images & Bicubic  & SC &K-SVD & ANR & SR-CNN & $\text{MAP}_{7 \times 7}$ \\
\hline \hline
baboon & 23.21/54.39 & 23.47/58.78 &23.52/58.99 & 23.56/59.90 & \textbf{23.60}/60.35 & 23.58/\textbf{61.50}\\
barbara & 26.25/75.31 & 26.39/76.33 &\textbf{26.76}/78.16 & 26.69/78.11 & 26.66/78.10 & 26.43/\textbf{78.30} \\
bridge & 24.40/64.83 & 24.82/69.21 & 25.02/69.64 & 25.01/70.12 & 25.07/70.50 & \textbf{25.13}/\textbf{71.91} \\
coastguard & 26.55/61.49 & 27.02/63.93 & 27.15/65.40 &27.08/65.77 & 27.20/65.88 & \textbf{27.25}/\textbf{67.04}\\
comic & 23.12/69.88 &23.90/75.57 & 23.96/75.48 & 24.04/76.06 & \textbf{24.39}/77.78 & 24.26/\textbf{78.36}\\
face & 32.82/79.84 & 33.11/80.11 & 33.53/81.95 & 33.62/82.32 & 33.58/82.14 & \textbf{33.70}/\textbf{82.91} \\
flowers & 27.23/80.13 & 28.25/82.97 & 28.43/83.75 &28.49/84.03 & \textbf{28.97}/84.75 & 28.84/\textbf{85.58}\\
foreman & 31.18/90.58 & 32.04/91.32 & 33.19/92.92 & 33.23/93.01 & 33.35/93.21 & \textbf{33.83}/\textbf{93.91} \\
lenna & 31.68/85.82 & 32.64/86.48 & 33.00/87.82 & 33.08/88.04 & \textbf{33.39}/88.27 & 33.31/\textbf{88.64} \\
man & 27.01/74.95 & 27.76/77.49 & 27.90/78.62 & 27.92/78.90 & \textbf{28.18}/79.40 & 28.15/\textbf{80.27}\\
monarch & 29.43/91.98 & 30.71/92.90 & 31.10/93.82 &31.09/93.77 & \textbf{32.39}/94.50 & 31.88/\textbf{94.77} \\
pepper & 32.39/86.98 & 33.32/86.69 & 34.07/88.63 & 33.82/88.51 & \textbf{34.35}/88.76 & 34.30/\textbf{89.15} \\
ppt3 & 23.71/87.46 & 24.98/89.18 & 25.23/91.17 & 25.03/90.25 & 26.02/91.94 & \textbf{26.42}/\textbf{93.52} \\
zebra & 26.63/79.42 & 27.95/82.59 & 28.49/84.09 & 28.43/84.24 & \textbf{28.87}/84.70 & 26.81/\textbf{85.42}\\
\hline \hline
\rowcolor[gray]{0.85} \textbf{average} & 27.54/77.36 & 28.31/79.54 & 28.67/80.75 
& 28.65/80.93 & \textbf{29.00}/81.45 & \textbf{28.99}/\textbf{82.23}\\
\hline
\hline
\textbf{Set5} images & Bicubic & SC &K-SVD & ANR & SR-CNN & 
$\text{MAP}_{7 \times 7}$ \\
\hline
baby	 &33.91/90.39 & 34.29/90.42 &35.08/92.11 & \textbf{35.13}/92.25& 35.01/92.10 & 35.10/\textbf{92.61}\\
bird & 32.58/92.56 & 34.11/93.91 &34.57/94.77 & 34.60/94.88 & 34.91/94.94 & \textbf{35.07}/\textbf{95.56}\\
butterfly & 24.04/82.16 &25.58/86.11& 25.94/87.70 & 25.90/87.17 & \textbf{27.58}/90.12& 26.79/\textbf{90.31} \\
head & 32.88/80.03 & 33.17/80.24 & 33.56/82.04 &33.63/82.39 & 33.55/82.15 & \textbf{33.72}/\textbf{82.99}\\
woman & 28.56/88.96 & 29.94/90.37 & 30.37/91.76 & 30.33/91.69 & \textbf{30.92}/92.36 & 30.79/\textbf{92.80}\\
\hline
\hline
\rowcolor[gray]{0.85} \textbf{average} & 30.39/86.82 & 31.42/88.21 
& 31.90/89.68 & 31.92/89.68 & \textbf{32.39}/90.33 & 32.29/\textbf{90.85}\\
\hline
\end{tabular}
\caption{Scale $\times 3$ performance in terms of PSNR and SSIM ($\times 100$) 
index on the Set14 and Set5 datasets. We compare the results of 
MAP based SR model (48 filters of size $7 \times 7$) to representative 
state-of-the-art SR methods: SC \cite{SCSR}, ANR 
\cite{ANR}, K-SVD \cite{SRKSVD} and SR-CNN \cite{SRCNN}. }\label{table:set14}
\vspace*{-0.75cm}
\end{table}

\begin{figure}[t!]
\vspace*{-0.3cm}
\centering
    \subfigure[Original/PSNR (-)]{\includegraphics[width=0.245\textwidth]{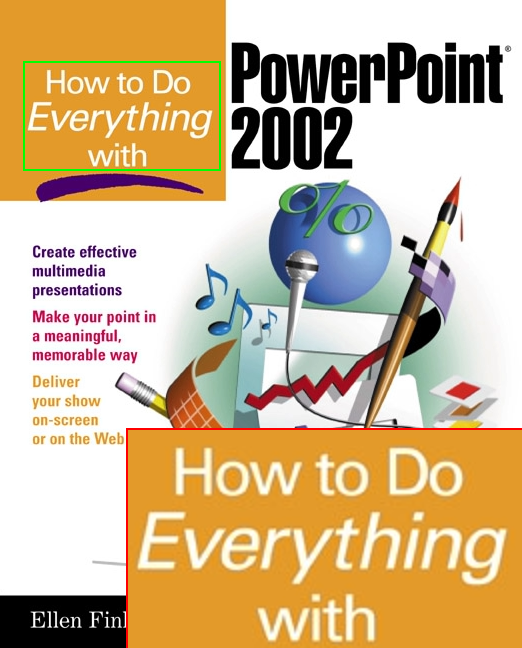}}\hfill
    \subfigure[Bicubic/23.71dB/87.46 (-)]{\includegraphics[width=0.245\textwidth]{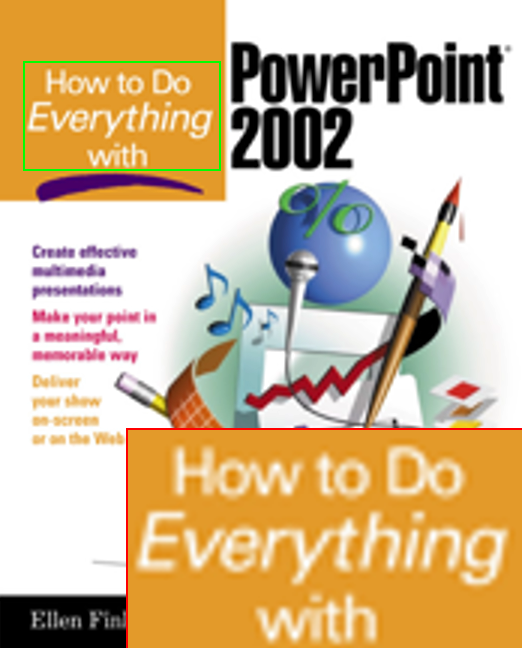}}\hfill
	\subfigure[SC/24.98dB/89.18 {(89.2s)}]{\includegraphics[width=0.245\textwidth]{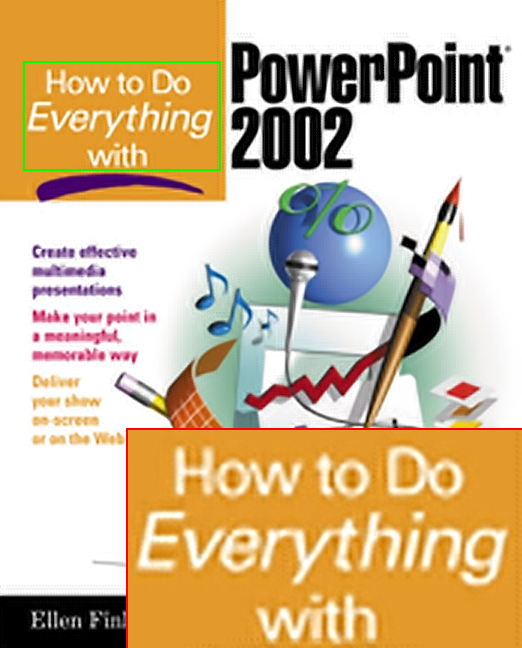}}\hfill
    \subfigure[ANR/25.03dB/90.25 {(1.1s)}]{\includegraphics[width=0.245\textwidth]{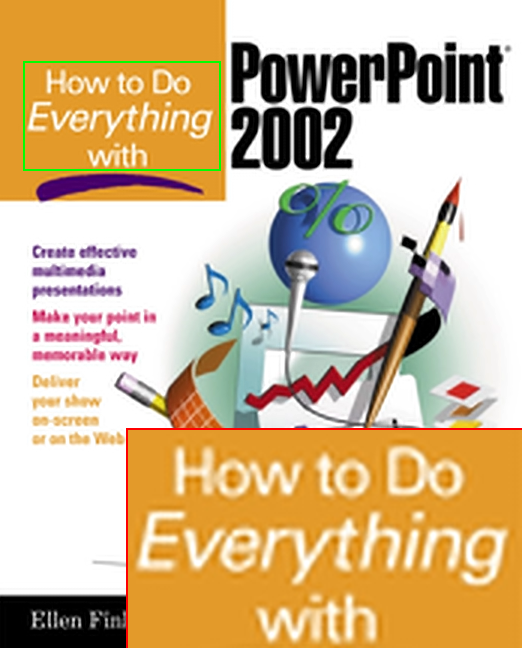}}\\
    \subfigure[K-SVD/25.23dB/91.17 {(4.5s)}]{\includegraphics[width=0.28\textwidth]{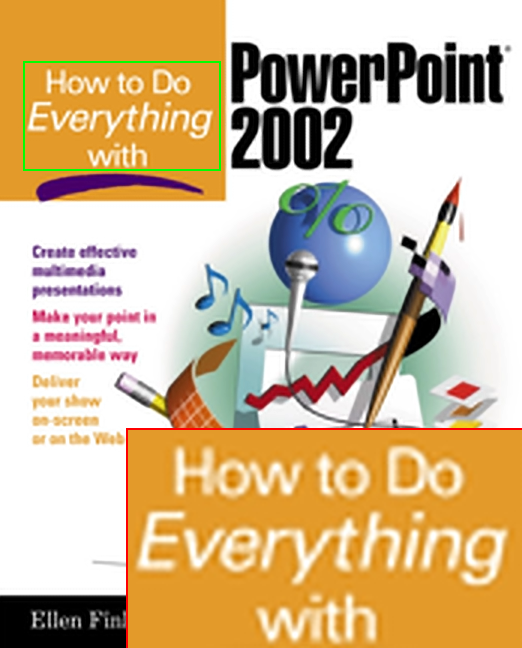}}
    \subfigure[SR-CNN/26.02dB/91.94 {(4.7s)}]{\includegraphics[width=0.28\textwidth]{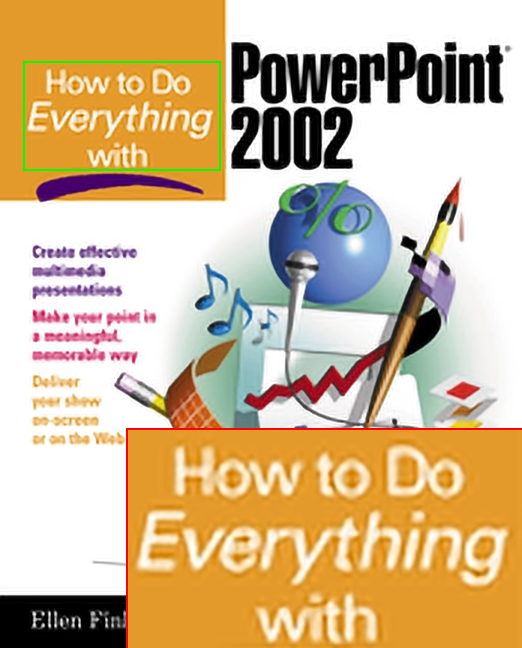}}
    \subfigure[$\text{MAP}_{7 \times 7}$/\textbf{26.42}dB/\textbf{93.52}
 {(45.3s)}]
{\includegraphics[width=0.28\textwidth]{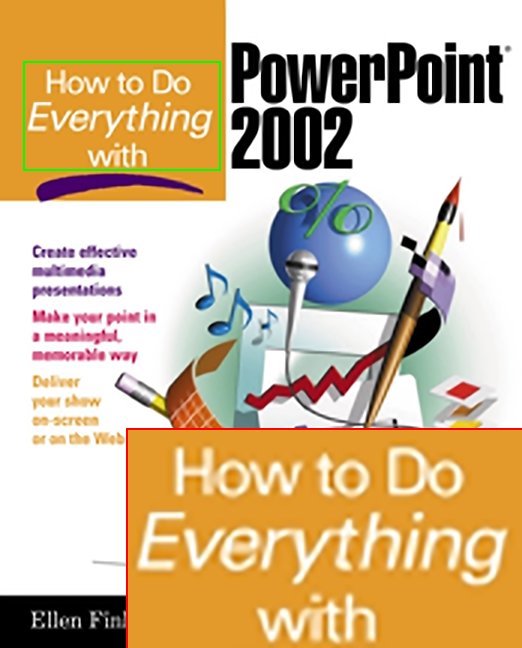}}\\
    \caption{Visual qualitative assessment and quantitative results for ``ppt3'' image from \textbf{Set14} with magnification $\times 3$ 
(best viewed on screen). 
Note the difference in the highlighted region. The corresponding run 
time is given in the bracket. }\label{SRppt}
\vspace*{-0.4cm}
\end{figure}

The SR results on \textbf{Set14} and \textbf{Set5} 
are summarized in Table \ref{table:set14}. We can see that the FoE based SR model with filters of size $7 \times 7$ 
achieves similar average PSNR as the SR-CNN method, and outperforms other competing algorithms. 
A visual example is shown in Figure \ref{SRppt}\footnote{Following \cite{SRCNN}, we only consider the 
luminance channel (in YCrCb color space) in our experiments. The two chrominance 
channels are directly upsampled using the bicubic interpolation for the purpose of display. }. 
In the highlighted region, one can see that our SR method achieve more 
clear edges than other approaches. 
In summary our $\text{MAP}_{7 \times 7}$ model obtains strongly competitive quality performance to current state-of-the-art SR methods. {
Furthermore, we also provide 
the run time of the exploited SR algorithms. Note that all the algorithms are run 
in Matlab, and therefore, the SR-CNN algorithm 
is slower than its C++ implementation shown in \cite{SRCNN}. 

In order to illustrate the convergence properties of the iPiano algorithm used to 
solve the MAP-based SR problem, in Figure \ref{energy} 
we present the energy curve associated with the SR task for the ``ppt3''image 
shown in Figure \ref{SRppt}.
\begin{figure}[t!]
\vspace*{-0.25cm}
\centering
    {\includegraphics[width=0.5\textwidth]{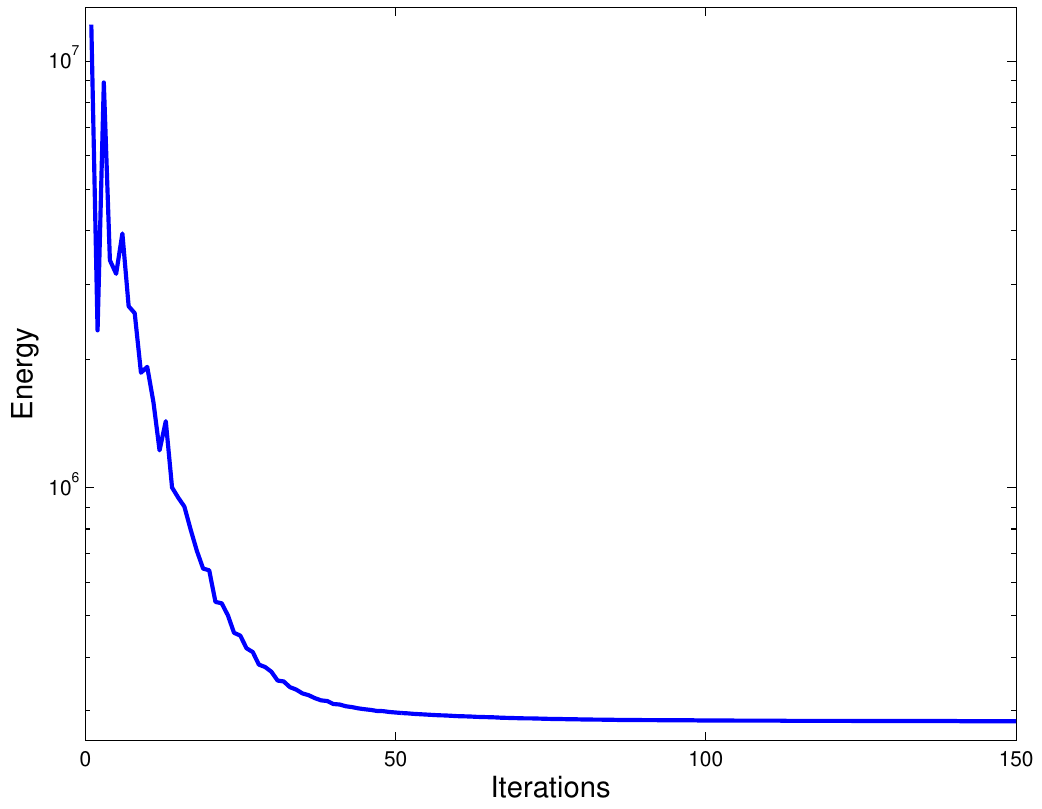}}
\vspace*{-0.3cm}
    \caption{Energy curve of the iPiano algorithm used to solve the SR problem 
for the image ``ppt3'' shown in Figure \ref{SRppt}.}
\label{energy}
\end{figure}
}

\section{Discussion and Conclusion}
In the context of higher-order MRF based models, 
it is generally true that the MAP estimate, which only seeks for the posterior mode, 
could not generally exploit the full potential offered by the 
probabilistic modeling, while the MMSE estimate, which directly draw samples from the probability model, should be more 
powerful. On the other hand, it is well-known that the sampling based MMSE estimation is very slow, making the corresponding 
methods hardly appealing for practical applications if one has to stick to the MMSE inference. 

In this paper, we have concentrated on the higher-order MRFs based SR problem, and evaluated the performance 
of the MAP estimate in inference. We found that the MAP estimate can work equally well compared to MMSE in the presence of 
the same FoE prior, despite of the non-convexity of the resulting optimization problem. We believe the reason is two-folds: 
first, the exploited iPiano algorithm which is an effective non-convex optimization algorithm, helps us reach the MAP mode in a 
short time; secondly, in practice one is not able to obtain an accurate solution for the MMSE estimate. 
In addition, we found that the performance of MAP estimate can be further 
boosted by using discriminatively trained FoE prior models. 
As a consequence, the resulting $\text{MAP}_{7 \times 7}$ model, which involves 
48 filters of size $7 \times 7$ can lead to strongly competitive results to very recent state-of-the-art SR methods. 
Therefore, concerning the higher-order MRFs based SR task, we suggest to exploit the MAP estimate for inference because 
there is no performance loss by using this simpler inference criterion while it has an obvious advantage of high efficiency.

Furthermore, it is notable to point out that the findings about the MAP estimate presented in this paper strengthen our arguments 
drawn based on the Gaussian denoising problem in our previous works \cite{ChenPRB13, ChenRP14}. 
We have show in \cite{ChenPRB13, ChenRP14} that the MAP-based denoising model 
with our discriminatively trained FoE prior leads to the best results among the MRF-based systems, including MMSE based models. 
Therefore, we believe that MAP-based denoising model does not perform well in previous works, \eg, 
\cite{gaocrpr2010, SamuelFoE} just because they have not obtained a good FoE prior well-suited for the MAP inference. 

In summary, we believe that in the context of higher-order MRF image prior based modeling for image restoration problems, 
it is a better choice to make use of the MAP estimate, together with the discriminatively trained FoE prior. 
\bibliographystyle{ieee}
\bibliography{references}

\end{document}